\pdfoutput=1

\documentclass{article}

\newcommand{\FF}{FilterForward\xspace}

\usepackage{microtype}
\usepackage{graphicx}
\usepackage{booktabs}  
\usepackage{subcaption}

\usepackage{hyperref}

\usepackage[accepted]{sysml2019}

\usepackage{url}
\makeatletter
\g@addto@macro{\UrlBreaks}{\UrlOrds}
\g@addto@macro{\UrlBreaks}{\do\/\do\a\do\b\do\c\do\d\do\e\do\f\do\g\do\h\do\i\do\j\do\k\do\l\do\m\do\n\do\o\do\p\do\q\do\r\do\s\do\t\do\u\do\v\do\w\do\x\do\y\do\z\do\A\do\B\do\C\do\D\do\E\do\F\do\G\do\H\do\I\do\J\do\K\do\L\do\M\do\N\do\O\do\P\do\Q\do\R\do\S\do\T\do\U\do\V\do\W\do\X\do\Y\do\Z\do\1\do\2\do\3\do\4\do\5\do\6\do\7\do\8\do\9\do\0\do\.}
\makeatother

\pdftrailerid{}  
\usepackage{soulutf8}
\soulregister\cite7
\soulregister\ref7
\soulregister\pageref7

\usepackage{xspace}

\usepackage{mathtools}

\usepackage{newtxtext,newtxmath}  

\usepackage{textcomp}

\renewcommand{\S}{Section}

\sysmltitlerunning{Scaling Video Analytics on Constrained Edge Nodes}

\begin{document}

\twocolumn[
\sysmltitle{Scaling Video Analytics on Constrained Edge Nodes}

\sysmlsetsymbol{equal}{*}

\begin{sysmlauthorlist}
  \sysmlauthor{Christopher Canel}{equal,cmu}
  \sysmlauthor{Thomas Kim}{equal,cmu}
  \sysmlauthor{Giulio Zhou}{cmu}
  \sysmlauthor{Conglong Li}{cmu}
  \sysmlauthor{Hyeontaek Lim}{cmu}
  \sysmlauthor{David G. Andersen}{cmu}
  \sysmlauthor{Michael Kaminsky}{intel}
  \sysmlauthor{Subramanya R. Dulloor}{thoughtspot}
\end{sysmlauthorlist}

%
\sysmlaffiliation{cmu}{Computer Science Department, School of Computer Science,
  Carnegie Mellon University, Pittsburgh, Pennsylvania, USA}
\sysmlaffiliation{intel}{Intel Labs, Pittsburgh, Pennsylvania, USA}
\sysmlaffiliation{thoughtspot}{ThoughtSpot, Palo Alto, California, USA}

\sysmlcorrespondingauthor{Christopher Canel}{ccanel@cmu.edu}

\sysmlkeywords{Edge, Video, Object Detection, Machine Learning, SysML}

\vskip 0.3in

\begin{abstract}

As video camera deployments continue to grow, the need to process large volumes
of real-time data strains wide area network infrastructure. When per-camera
bandwidth is limited, it is infeasible for applications such as traffic
monitoring and pedestrian tracking to offload high-quality video streams to a
datacenter. This paper presents \FF, a new edge-to-cloud system that enables
datacenter-based applications to process content from thousands of cameras by
installing lightweight edge filters that backhaul only relevant video
frames. \FF introduces fast and expressive per-application ``microclassifiers''
that share computation to simultaneously detect dozens of events on
computationally constrained edge nodes. Only matching events are transmitted to
the cloud. Evaluation on two real-world camera feed datasets shows that \FF
reduces bandwidth use by an order of magnitude while improving computational
efficiency and event detection accuracy for challenging video content.

  This paper is an extended version of~\cite{canel:sysml2019}.
\end{abstract}
]  



\printAffiliationsAndNotice{\sysmlEqualContribution} 


\section{Introduction}
\label{sec:intro}

Video camera deployments in urban areas are ubiquitous: in malls, offices, and
homes, and on streets, cars, and people. Almost 100 million networked
surveillance cameras were purchased worldwide in 2017~\cite{www-ihs}. Machine
learning--based analytics on real-time streams collected by these cameras, such
as traffic monitoring, customer tracking, and event detection, promise
breakthroughs in efficiency and safety. However, tens of thousands of always-on
cameras installed in a modern city collectively generate hundreds of gigabits of
data every second, overloading shared network infrastructure.  This problem is
worse for wirelessly and cellularly--connected nodes and areas outside of
infrastructure-rich metropolitan centers~\cite{www-fcc-rural-internet}, as they
often have more constrained
networks~\cite{www-google-wireless-internet,broadband-commission:2017}.
Moreover, the infeasibility of uploading streaming video is at odds with the
growing complexity of video analytics applications, which are designed to run in
datacenters. This paper addresses the question of how to overcome this network
bottleneck and offload large volumes of data from a distributed camera
deployment in real time to a datacenter for further processing.

Deployment proliferation, combined with increasing camera resolution,
necessitates an edge-based filtering approach that is parsimonious with limited
bandwidth. We present \emph{\FF}, a system that offers the benefits of both edge
computing and datacenter-centric approaches to wide-area video processing.
Using edge-compute resources collocated with the cameras, \FF identifies the
video sequences that are most relevant to datacenter applications
(``filtering'') and offloads only that data for further analysis
(``forwarding''). In this way, \FF supports near-real-time processing running in
datacenters while limiting the use of low-bandwidth wide area network links.

\FF is designed for scenarios meeting two key assumptions, which hold for some,
though certainly not all, applications. First, relevant events are rare. There
is bandwidth to be saved by transmitting only relevant data. Second, datacenter
applications require high-quality video data to complete their tasks. This
precludes solutions such as heavily compressing streams or reducing their
spatial (frame dimensions) or temporal (frame frequency) resolutions.

In the \FF model, datacenter applications express interest in specific types of
visual content (e.g., ``send me sequences containing dogs''). Each application
installs on the edge a set of small neural networks called
\emph{microclassifiers (MCs)} that perform binary classification on each
incoming frame to determine whether an interesting state is occurring.
Typically, an interesting state is described in terms of the presence of a
certain object. Each MC is trained offline by an application developer. At
runtime, frame-level classification results are smoothed to determine the start
and end points of ``events'' during which the interesting state occurred. Events
are re-encoded and streamed to the datacenter.

\FF scales to multiple independent applications (e.g., ``find dogs and find
bicycles'') by evaluating many MCs in parallel. Optimizing this multi-tenancy is
\FF's key contribution. Instead of designing the MCs to operate on raw pixels,
\FF draws inspiration from modern object detectors and uses a shared base deep
neural network (DNN) to extract general features from each frame. All MCs
operate on the activations from the base DNN, but they may draw from different
layers. This amortizes the expensive task of pixel processing across all of the
MCs, allowing \FF to execute tens of concurrent MCs using the CPU power
available in a small form factor edge node. The base DNN is an expensive
per-frame, upfront overhead, but it enables a significant performance
improvement once the number of concurrent MCs passes a break-even point.

Our evaluation using two real-world camera feed datasets demonstrates that, for
applications meeting \FF's requirements (operating with severe bandwidth
constraints and requiring high-fidelity data), our architecture uses an order of
magnitude less bandwidth than standard compression techniques
(Section~\ref{sec:eval-bw}). Furthermore, \FF is computationally efficient,
surpassing the frame rate of existing lightweight filters~\cite{kang:vldb2017}
when more than $3-4$ MCs run together and achieving up to $6.1\times$ higher
throughput with 50 concurrent MCs (Section~\ref{sec:eval-e2e}). Finally, MCs are
up to $1.3\times$ more accurate than pixel-based DNN filters used in prior work
while having up to a $23\times$ lower marginal cost (Section~\ref{sec:eval-mc}).

\FF is open source at
\href{http://www.github.com/viscloud/ff}{github.com/viscloud/ff}.


\section{Background and Challenges}
\label{sec:back-chall}

This section provides an overview of video analytics before delving into the key
challenges introduced by a large-scale camera deployment.

\subsection{Video Analytics}
\label{sec:back-chall-video-analytics}

Typical video analytics primitives include: \emph{Image classification}
categorizes a whole frame based on its most dominant features (e.g., ``This is
an image of an intersection.''). \emph{Object detection} finds interesting
objects that may occupy only a small portion of the view and categorizes them
(e.g., ``This rectangle defines a region containing a car.''). \emph{Object
  tracking} aims to label each object's location across multiple frames (e.g.,
``This path plots the progress of pedestrian $A$ crossing the road.''). These
and other primitives form the basis of more advanced analyses, such as traffic
monitoring, pedestrian action understanding, and hazard detection.

Video analytics workloads entail extensive computation on large amounts of data
(e.g., a $1920\times1080$~pixel stream at 30 frames per second (fps) is
$\approx1.5$~Gb/s when decompressed). Accomplishing video analytics at scale
requires abundant compute, memory, and storage resources, so existing systems
often perform this processing in the cloud, using
GPUs~\cite{kang:vldb2017,Zhang:nsdi2017}.

\subsection{Edge-to-cloud Challenges}
\label{sec:back-chall-chall}

The scenarios that motivate \FF include remote ``Internet of Things'' monitoring
and ``smart city'' deployments of tens or hundreds of thousands of wide-angle,
fixed-view cameras. In this section, we describe three key challenges presented
by this use case.

\subsubsection{Limited Bandwidth}
\label{sec:back-chall-bw}

Running video analytics by streaming all video to the cloud conflicts with the
bandwidth constraints of some deployments, which preclude uploading all camera
data. Each camera's uplink bandwidth is limited, both by the physical
constraints of modern wide area network infrastructure and the monetary cost of
operating a widespread camera deployment. Specifically, we consider large-scale
deployments where each camera receives a bandwidth allocation of a few hundred
kilobits per second, or less~\cite{www-rfp-camera-system}. For comparison, a
\emph{low-quality} H.264-encoded 1080p ($1920\times1080$ pixels) stream is
approximately 2~Mb/s, an order of magnitude greater than our available uplink
bandwidth. Yet, such low-quality data is often insufficient to perform accurate
analyses. Modern 4K ($3840\times2160$ pixels) cameras produce up to 30-40~Mb/s,
two orders of magnitude beyond the uplink bandwidth, and this gap will only
expand as 8K ($7680\times4320$ pixels) cameras become more common. As a concrete
example, we built an off-campus deployment where cameras are mounted next to
traffic lights at an intersection. The local Internet service provider charges
\$400 per month for a single 35~Mb/s uplink, creating a strong economic
incentive for us to share that bandwidth between as many cameras as possible
(currently, eight 4K cameras share each uplink).

This bandwidth gap, exacerbated by the requirement for high-quality data,
necessitates an edge-based decision about which frames to send to the
datacenter. \FF answers this challenge with semantic filtering that uploads only
the frames that are relevant to applications.

\begin{figure*}[t]
  \centering \includegraphics[width=0.7\textwidth]{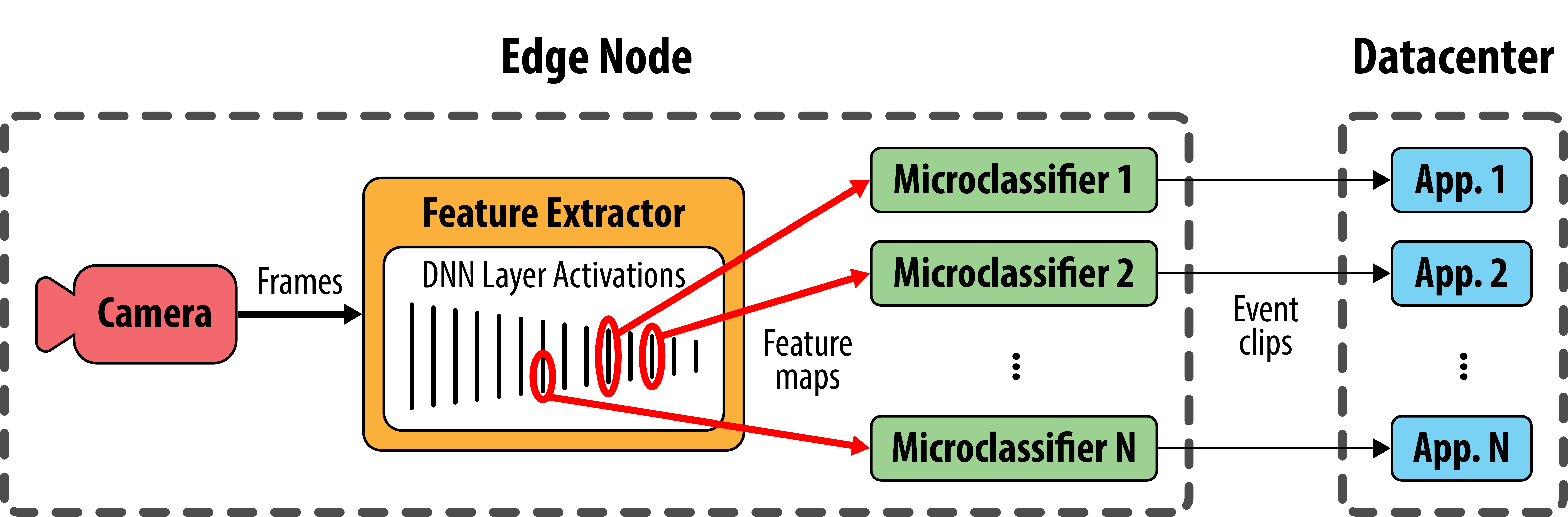}
  \caption{The \FF architecture.}
  \vskip -1em
  \label{fig:ff_arch}
\end{figure*}

\subsubsection{Real-world Video Streams}
\label{sec:back-chall-real-world}

In many surveillance deployments, cameras are mounted high on buildings or light
posts and fitted with wide-angle lenses that capture broad views of the
surrounding area. Interesting objects (e.g., pedestrians, license plates,
parcels, animals, etc.) occupy a small portion of the frame. This poses a
challenge for the video analytics primitives discussed in
Section~\ref{sec:back-chall-video-analytics}. Image classification maps the
entire image to a single category, so an urban viewpoint would always be labeled
``street'' or ``traffic,'' which is of limited use. Object detection and
tracking are designed to pick individual objects out of a frame but often
operate on low-resolution images (e.g., as small as $300\times300$
pixels~\cite{liu2016ssd}). Aggressive downsampling of a wide-angle image causes
details such as license plates and distant people to disappear. To detect these
fine-grained details, \FF introduces microclassifiers that process
high-resolution images on the edge, avoiding quality degradation caused by the
decimation required to meet bandwidth constraints.

\subsubsection{Scalable Multi-tenancy}
\label{sec:back-chall-scalability}

In real-world deployments, cameras observe scenes containing diverse objects and
activities. A single camera may record pedestrians walking down the sidewalk,
vehicles stopped at a traffic light, and shoppers entering stores; all while
capturing the current weather, the quantity of leaves on the trees, and whether
there is snow on the roads. Different applications are simultaneously interested
in all of this information, and more. Therefore, any edge-filtering approach
must scale to multiple cloud applications focused on disjoint regions of the
frame in parallel.

Given edge nodes' limited compute resources, scaling to multiple applications
naturally poses a performance challenge. A na\"ive approach to handling $N$
applications is to run $N$ full DNNs concurrently. However, even relatively
lightweight DNNs are costly. In our experience, on a modest
Intel\textsuperscript{\textregistered{}} CPU (not GPU),
MobileNet~\cite{DBLP:journals/corr/HowardZCKWWAA17} runs at approximately 15~fps
for $512\times512$~pixel images while consuming more than 1~GB of memory. Even
lightweight DNNs have high resource requirements, precluding execution of more
than a handful in real time on an edge node. Therefore, to achieve scalability,
\FF's lightweight MCs simplify per-application processing while the base DNN
shares redundant computation between applications.

The rest of this paper describes how \FF addresses the above challenges:
semantic filtering decimates video streams to meet bandwidth constraints, novel
microclassifier architectures detect fine-grained details in wide-angle video,
and computation reuse enables scalable multi-tenancy.


\section{Designing \FF}
\label{sec:design}

\FF (FF) is a novel video analytics platform that reuses computation to provide
highly accurate, multi-tenant video filtering for bandwidth-constrained edge
nodes. Purely edge-based approaches constrain applications to the static compute
and storage resources of field installations, while datacenter-only analytics
necessitate heavily compressing the video for transport. FF offers applications
the flexibility of splitting their work between the edge and the cloud, taking
advantage of high-fidelity data at the edge to make relevant video sequences
available in the cloud.

This section describes the architecture of FF's two main components, the feature
extractor and microclassifiers, and explains how they address the three
challenges described in Section~\ref{sec:back-chall-chall}: meeting bandwidth
constraints, detecting subtle details in real-world video, and supporting many
concurrent applications. The system architecture is shown in
Figure~\ref{fig:ff_arch}.

\subsection{Generating Features}
\label{sec:design-fvgen}

In FF, microclassifiers reuse computation by taking as input \emph{feature maps}
produced from the intermediate results (\emph{activations}) of a single
reference DNN, which we refer to as the \emph{base DNN}\@. The component that
evaluates the base DNN and produces feature maps is called the \emph{feature
  extractor}.

As prior work observes~\cite{sharghi:cvpr2017, yeung:cvpr2016}, activations
capture information that humans intuitively desire to extract from images, such
as the presence and number of objects in a scene, and outperform handcrafted
low-level features~\cite{razavian:cvprw2014, ng:cvprw2015, babenko:iccv2015}.
The activations of the first layers of a DNN (often simple convolutional filters
such as edge detectors) are still visually recognizable. Later activations
represent high-level concepts (e.g., ``eye,'' ``fur,'' etc.). Processing feature
maps created from these activations has been used successfully for tasks such as
object region proposals, segmentation, and tracking~\cite{ren2015faster,
  hariharan2015hypercolumns, ma2015hierarchical, bertinetto2016fully}, as well
as action classification~\cite{sharma2015action}.

For our evaluation, we use the
MobileNet~\cite{DBLP:journals/corr/HowardZCKWWAA17} architecture trained on
ImageNet~\cite{Russakovsky:ijcv2015} as the base DNN\@. MobileNet offers a
balance between accuracy and computational demand that is appropriate for
constrained edge nodes. We use the 32-bit (unquantized) version of the
network. Picking an appropriate base DNN is, of course, a moving target, and we
do not view the selection of a \emph{specific} network as a contribution of this
work. Evaluating the robustness of our filtering algorithm to different base
DNNs is left for future research.

The feature extractor plays a crucial role in addressing the challenge of
detecting small objects in real-world surveillance video
(Section~\ref{sec:back-chall-real-world}). Instead of drastically shrinking
incoming frames as is typical in ML-based video analytics, \FF examines
full-resolution frames. For our evaluation, the full resolution is either
$1920\times1080$ or $2048\times850$~pixels (Section~\ref{sec:eval-datasets}),
which represent $41.3\times$ and $34.7\times$ increases in total input data,
respectively, versus the typical MobileNet input size of
$224\times224$~pixels. By operating on high-resolution frames, small content
such as distant pedestrians, make and model--specific automobile details, and
faces are captured in greater detail.

However, processing drastically more pixels imposes a significant computation
overhead, as the work done by each layer of the base DNN increases. Although
feature extraction is the most computationally intensive phase of \FF, its
results are reused by all of the MCs, amortizing the per-frame, upfront overhead
once the number of MCs passes a break-even point. This computation reuse is the
key to achieving scalable multi-tenancy, a major challenge for real-world
surveillance deployments (Section~\ref{sec:back-chall-scalability}).  Ongoing
architectural improvements in off-the-shelf feature extraction networks, as well
as advances in hardware accelerators~\cite{DBLP:journals/corr/JouppiYPPABBBBB17,
  www-apple-neural-engine, www-intel-movidius, www-microsoft-project-brainwave},
will continue to reduce \FF's computational overhead.

We evaluate the base DNN's computation overhead in
Section~\ref{sec:eval-e2e}. Ultimately, the feature maps generated by the base
DNN underpin FF's accuracy and scalability achievements.

\subsection{Finding Relevant Frames Using Microclassifiers}
\label{sec:design-mcs}

Microclassifiers are lightweight binary classification neural networks that take
as input feature maps extracted by the base DNN and output the probability that
a frame is relevant to a particular application. An edge node can run many MCs
on a single camera stream, or fewer MCs on several streams.

An application developer chooses an MC architecture (we present several
possibilities in Section~\ref{sec:design-mc-archs}) and trains it offline to
detect the application's desired content. To deploy an MC, the developer
supplies the network weights and architecture specification along with the name
of the base DNN layer (and, optionally, a crop thereof) to use as input.

Each microclassifier can pull feature maps from any layer of the base DNN,
enabling \FF to support different types of tasks
(Section~\ref{sec:back-chall-scalability}). Section~\ref{sec:design-mc-inputs}
discusses the layer selection process.

Furthermore, each MC can optionally crop its feature map, thus focusing on a
certain portion of the frame. Selecting a static subregion of the field of view
(for each MC) helps specialize \FF to wide-angle surveillance video
(Section~\ref{sec:back-chall-real-world}) because some applications are only
interested in particular regions. One benefit is that this reduces an MC's
computation load proportional to the decrease in its input size. Additionally,
constraining an MC's spatial scope increases accuracy (for certain applications)
for two reasons: (1) The MC must only consider the relevant region of the frame,
and (2) by cropping, important objects become more prominent. A key insight is
that by cropping feature maps instead of raw pixels, \FF retains its ability to
simultaneously support MCs interested in different regions, a key scalability
requirement. I.e., cropping an MC's feature map to refine its spatial scope is
an optional local optimization that each MC performs independently.

Dropping irrelevant frames is crucial to limiting bandwidth use, \FF's primary
objective (Section~\ref{sec:back-chall-bw}). Ideally, an MC will identify all of
the frames that an application needs to process in the cloud while rejecting a
large fraction of unimportant frames. The redundancy inherent in video provides
a safety margin for false negatives. False positives are particularly harmful
because they consume upload bandwidth with irrelevant data.

In the background, edge nodes record the original video stream to disk so that
datacenter applications can demand-fetch additional video (e.g., context
segments surrounding a matched segment) from the edge nodes' local storage.

As discussed in Section~\ref{sec:design-fvgen}, sharing computation between MCs
via the base DNN is \FF's solution to the multi-tenancy demands of real-world
surveillance deployments (Section~\ref{sec:back-chall-scalability}). We show in
Sections~\ref{sec:eval-e2e} and \ref{sec:eval-mc} that operating on feature maps
instead of raw pixels provides microclassifiers with competitive accuracy while
reducing marginal compute cost by an order of magnitude.

\subsection{Microclassifier Architectures}
\label{sec:design-mc-archs}

\begin{figure*}[t]
  \centering
  \begin{subfigure}[c]{0.45\textwidth}
    \centering \includegraphics[height=0.75in]{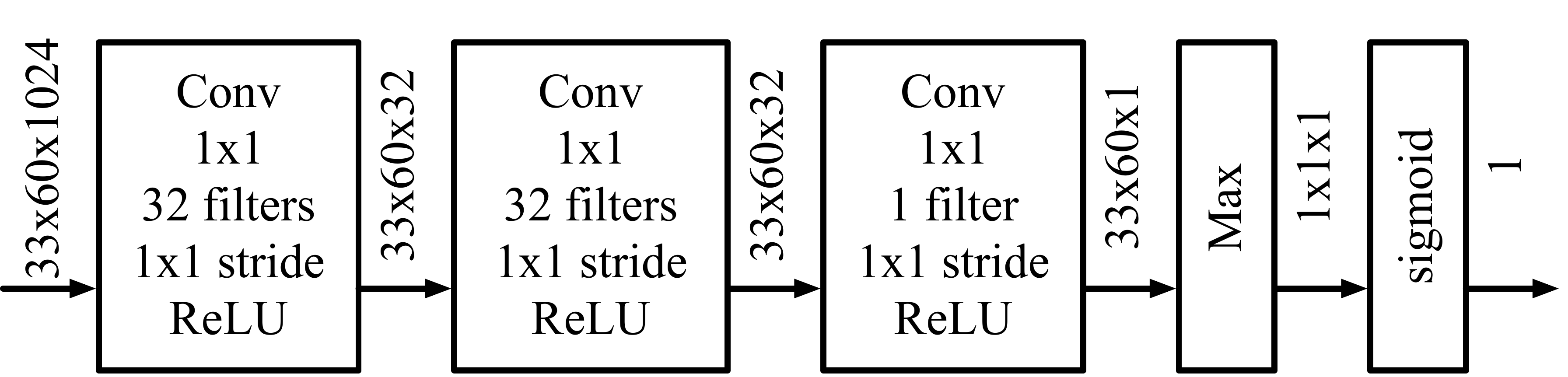}
    \caption{Full-frame object detector}
    \label{fig:objdet_arch}
  \end{subfigure}
  \hspace{1cm} ~
  \begin{subfigure}[c]{0.45\textwidth}
    \centering \includegraphics[height=0.75in]{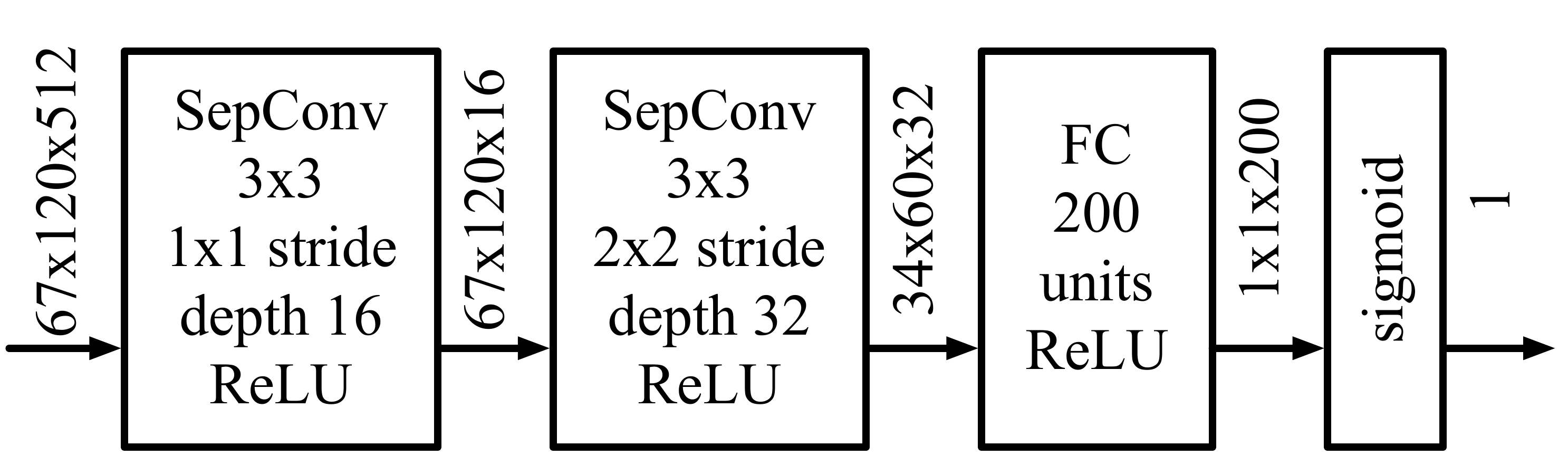}
    \caption{Localized binary classifier}
    \label{fig:spatial_arch}
  \end{subfigure}\qquad

  \begin{subfigure}[c]{0.8\textwidth}
    \centering \includegraphics[height=0.75in]{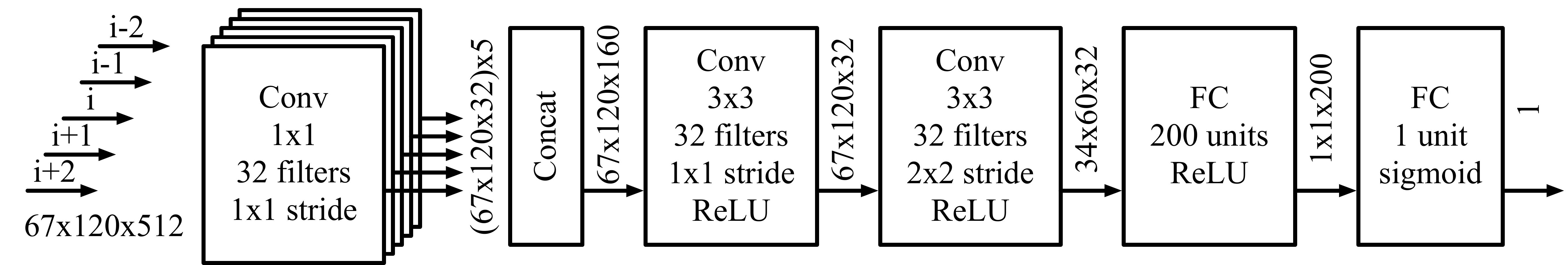}
    \caption{Windowed, localized binary classifier}
    \label{fig:window5_arch}
  \end{subfigure}

  \caption{Three microclassifier architectures. The dimensions quoted here
    correspond to a $1920\times1080$~pixel video (no spatial cropping).}
  \vskip -1em
  \label{fig:mc_archs}
\end{figure*}

As discussed in Section~\ref{sec:back-chall-real-world}, off-the-shelf
classifiers and detectors perform poorly on wide-angle surveillance video
because the objects of interest are often small. We propose three custom MC
architectures, shown in Figure~\ref{fig:mc_archs}, that solve this challenge in
different ways. In addition to the base DNN processing full-resolution frames,
two important microclassifier features that help achieve high accuracy on
surveillance video are: (1) MCs operate on activations from whichever base DNN
layer, and therefore whichever granularity of features, is most appropriate for
their task, while (2) optionally cropping away irrelevant regions of the
frame. Both of these capabilities help achieve high accuracy on real-world data,
as evaluated in Section~\ref{sec:eval-mc}.

\subsubsection{Full-frame Object Detector (Figure~\ref{fig:objdet_arch})}
\label{sec:design-ffod}

Modeled after sliding window--style object detectors such as
SSD~\cite{liu2016ssd} and Faster R-CNN~\cite{ren2015faster}, the full-frame
object detector MC applies a small binary classification DNN at each location in
a convolutional layer feature map and then aggregates the detections to make a
global prediction. This is achieved by using multiple layers of $1\times1$
convolutions and then applying a \texttt{max} operator over the grid of logits
(signifying looking for $\geq 1$ objects). This model is specifically designed
for pattern matching queries, with an implicit assumption of translational
invariance (i.e., the model runs the same template matcher everywhere), and is
well-suited to processing entire wide-angle frames.

\subsubsection{Localized Binary Classifier (Figure~\ref{fig:spatial_arch})}
\label{sec:design-lbc}

The localized binary classifier MC is a lightweight convolutional neural network
(CNN) that processes spatially cropped feature maps. Consisting of two separable
convolutions and a fully-connected layer, this architecture is designed to
detect prominent objects within a localized region (i.e., like zooming in to a
region of the frame).

\subsubsection{Windowed, Localized Binary Classifier
  (Figure~\ref{fig:window5_arch})}
\label{sec:design-wlbc}

This architecture extends the localized binary classifier MC to incorporate
nearby temporal context, improving per-frame accuracy.  The user specifies a
temporal window of $W$ frames. Given the convolutional feature maps for a
symmetric $W$-sized window centered at frame $F$, the windowed, localized binary
classifier MC first applies a $1\times1$ convolution to each frame's feature
map, then depthwise-concatenates the resulting activations and applies a CNN to
predict whether frame $F$ is interesting. This setup allows the MC to pick up on
motion cues in the scene, which helps achieve higher accuracies on tasks where
objects are constantly moving. The initial single-frame $1\times1$ convolution
significantly reduces the size of the input feature map, making this larger
architecture computationally tractable on edge node hardware. As an
optimization, the $1\times1$ convolutions are only computed once, and their
outputs are buffered and reused by subsequent windows, eliminating redundant
computation.

\subsection{Choosing Microclassifier Inputs}
\label{sec:design-mc-inputs}

Choosing which base DNN layer to use as input to each microclassifier is
critical to their accuracies. The layers of a CNN feature hierarchy offer a
tradeoff between spatial localization and semantic information. Too late a layer
may not be able to observe small details (because they have been subsumed by
global semantic classification). Too early a layer could be computationally
expensive due to the large size of early layer activations and the amount of
processing still required to transform low-level features into a classification.

As a baseline, we hand-select a layer, and optionally a crop region, based on
two heuristics. First, for the layer, we try to match the typical size of the
object class we were detecting. For example, to find pedestrians in a
$1920\times1080$~pixel video where the average height of a human is 40~pixels,
we choose the first layer at which a roughly 20:1--50:1 spatial reduction has
occurred. In our evaluation, the microclassifiers extract feature maps from the
following MobileNet layers: The full-frame object detector uses the penultimate
convolutional layer (\emph{conv5\_6/sep}) and the localized and windowed,
localized binary classifiers use a convolution layer from the middle of the
network (\emph{conv4\_2/sep}). Their names are specific to the version of
MobileNet that we use~\cite{www-mobilenet-caffe-port-cdwat}. Second, we choose
the optional crop region based on the region of interest for the application,
such as the crosswalks when detecting people (Section~\ref{sec:eval-datasets}).

In the prototype version of FF, each MC pulls features from a single base DNN
layer and we constrain the feature crops to be rectangular. Combining features
from multiple layers and experimenting with free-form and discontiguous crop
regions, as well as automating the selection of these parameters, are
interesting challenges for future work.

\subsection{From Per-frame Classifications to Events}
\label{sec:design-events}

A microclassifier outputs binary per-frame classifications (i.e., is this frame
relevant or not?), which \FF then smooths into event detections. First, each
MC's results for $N$ consecutive frames are accumulated into a window. Then, to
mask spurious misclassifications, we apply $K$-Voting to this window, treating
the middle frame as a detection if at least $K$ of the $N$ frames in the window
are positive detections. For our evaluation, we conservatively set $N = 5$ and
$K = 2$, which provides fairly aggressive false negative mitigation at the
expense of potential false positives. The resulting smoothed, per-frame labels
are fed into a transition detector that considers each contiguous segment of
positively-classified frames to be a unique event. Each event is assigned an
MC-specific, monotonically increasing, unique ID, which is stored in each
frame's metadata. These IDs are used by applications to determine the event
boundaries.

A single frame may be classified as part of an interesting event by multiple
MCs. For example, if frame $F$ is part of event $X$ for MC $A$ and event $Y$ for
MC $B$, then $F$'s internal metadata will contain the mapping
$(A \rightarrow X; B \rightarrow Y)$, indicating that it is part of multiple
events. As for the frames themselves, they are re-encoded using H.264 at a
user-configured bitrate and streamed back to the datacenter. The application
developer specifies a bitrate that is sufficiently high for their tasks (the
implications of this parameter are discussed in Section~\ref{sec:eval-bw}).


\section{Evaluation}
\label{sec:eval}

This section evaluates how \FF addresses the three challenges in
Section~\ref{sec:back-chall}: limited bandwidth, real-world video streams, and
scalable multi-tenancy. We begin by defining our datasets and accuracy metric,
then demonstrate that, for applications looking for rare events, \FF
significantly reduces bandwidth use. We show that \FF achieves a high frame rate
on commodity hardware while maintaining high accuracy on two event detection
tasks despite having a lower marginal cost than existing techniques.

\subsection{Real-world Datasets}
\label{sec:eval-datasets}

\begin{figure}[t]
  \centering

  \begin{subfigure}[c]{\columnwidth}
    \centering
    $\vcenter{\hbox{\includegraphics[width=0.45\columnwidth]{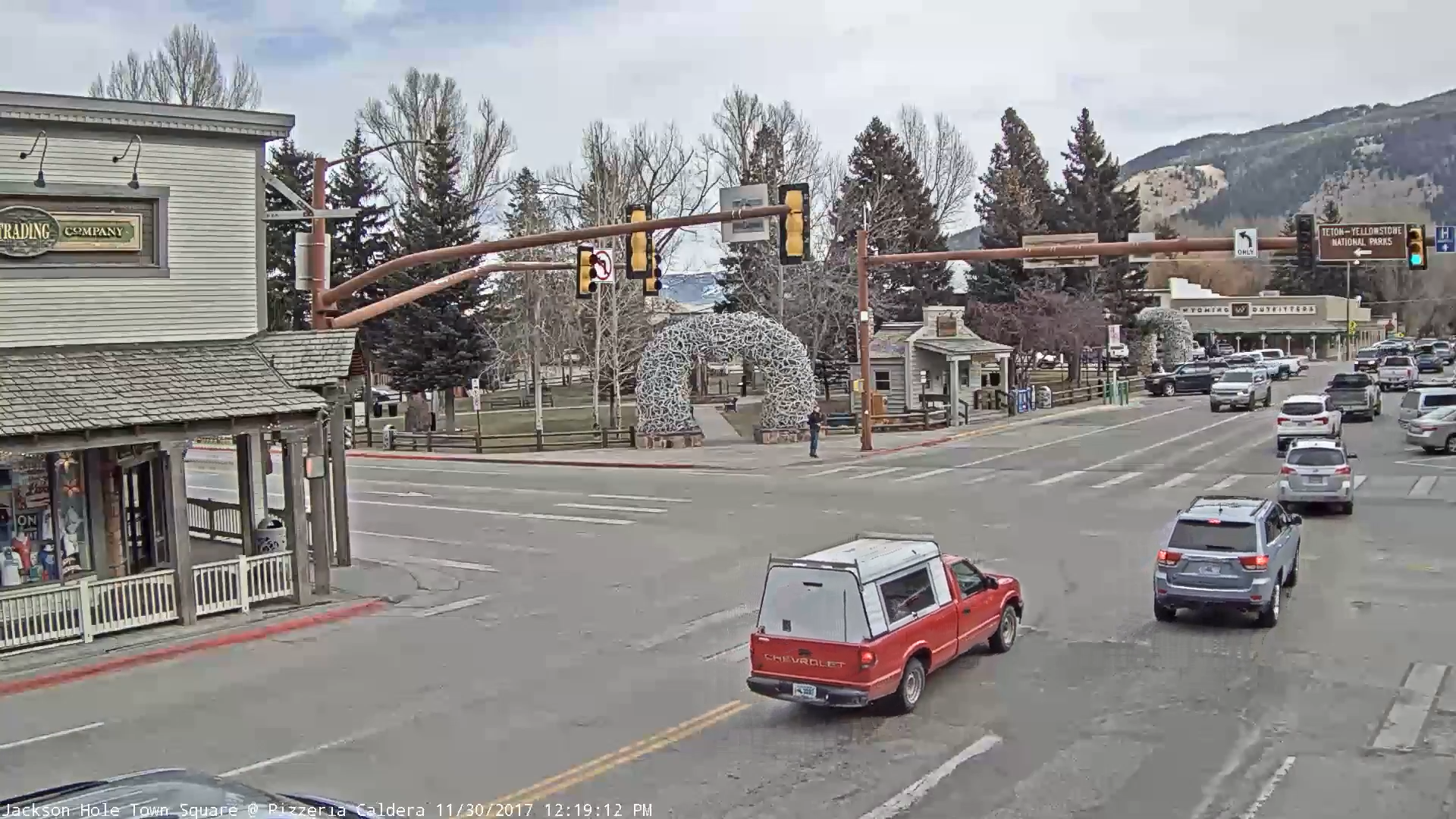}}}$
    \hspace*{0.025\columnwidth}
    $\vcenter{\hbox{\includegraphics[width=0.45\columnwidth]{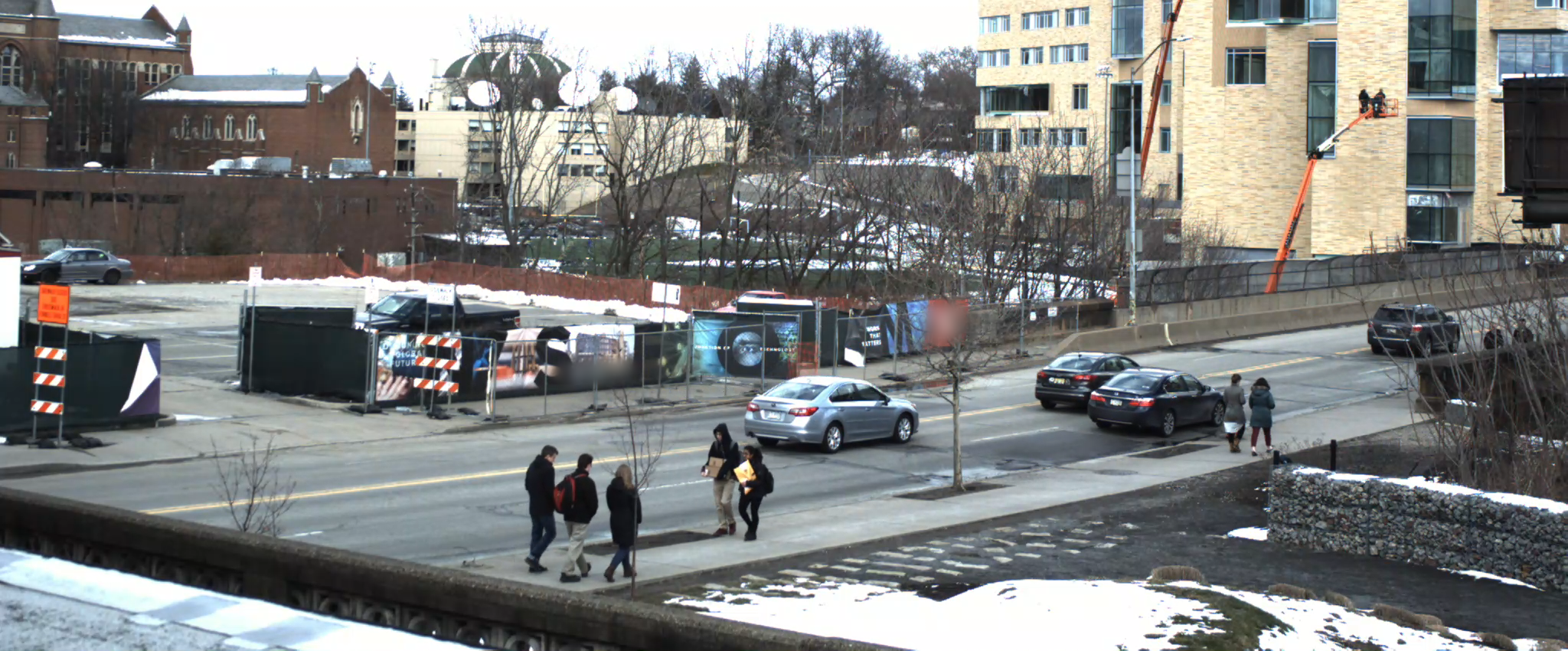}}}$

    \caption{The \emph{Jackson} and \emph{Roadway} datasets, respectively.}
    \label{fig:dataset-imgs}
  \end{subfigure}
  \vskip 0.15in

  \begin{subfigure}[c]{\columnwidth}
    \footnotesize \centering
    \caption{Dataset details.}

    \begin{center}
      \begin{small}
        \begin{tabular}{lcc}
          \textbf{Attribute} & \textbf{\emph{Jackson}} & \textbf{\emph{Roadway}} \\
          \midrule
          Resolution & $1920\times1080$~pixels & $2048\times850$~pixels \\
          Frame rate & 15~fps & 15~fps \\
          Frames & 600,000 & 324,009 \\
          Task & \emph{Pedestrian} & \emph{People with red} \\
          Event frames & 95,238 & 71,296 \\
          Unique events & 506 & 326 \\
        \end{tabular}
      \end{small}
    \end{center}
    \label{tbl:dataset-details}
  \end{subfigure}
  \vskip 0.15in

  \begin{subfigure}[c]{\columnwidth}
    \footnotesize \centering
    \caption{Rectangular pixel regions that correspond to the tasks' optional
      spatial crops. Note that in \FF, the feature maps are cropped, not the raw
      pixels.}

    \begin{center}
      \begin{small}
        \begin{tabular}{lcc}
          \textbf{Task} & \textbf{Upper left corner} & \textbf{Lower right corner} \\
          \midrule
          \emph{Pedestrian} & $(0, 539)$ & $(1919, 1079)$ \\
          \emph{People with red} & $(0, 315)$ & $(2047, 819)$ \\
        \end{tabular}
      \end{small}
    \end{center}
    \label{tbl:crop-details}
  \end{subfigure}

  \caption{Real-world evaluation videos and tasks.}
  \vskip -1em
  \label{fig:datasets}
\end{figure}

We evaluate using two datasets (Figure~\ref{fig:datasets}) showing scenes that
are representative of the real-world surveillance deployments that \FF
targets. The first dataset consists of video captured from a traffic camera
deployment in Jackson Hole, Wyoming (the \emph{Jackson} dataset). We collected
two six-hour videos from two consecutive days, between 10~AM and 4~PM. Then, we
annotated the twelve hours of data with labels for when pedestrians appear in
the crosswalks (the \emph{Pedestrian} task). This task allows us to demonstrate
the spatial selectivity of our microclassifiers in a way that is hopefully
relevant to future traffic monitoring applications. E.g., combined with a simple
traffic light classifier, a user could craft composite queries to detect
jaywalkers.

In addition, we collected a second dataset from a higher-quality camera in our
own urban deployment, consisting of two three-hour videos of a city street (the
\emph{Roadway} dataset) captured back-to-back during the middle of the day. We
annotated the six hours of data with labels for when passing pedestrians are
wearing red articles of clothing or carrying red parcels (the \emph{People with
  red} task). For both datasets, the first video is used for training and the
second for testing.

Table~\ref{tbl:crop-details} details the pixel regions corresponding to these
tasks' optional spatial crops.  For the \emph{Pedestrian} task, we select the
bottom half of the frame, as the trees and sky are unnecessary. For the
\emph{People with red} task, we select the street and sidewalk area ($59\%$ of
the frame). \FF crops the feature maps produced by the base DNN, not the
original pixels, so the coordinates in Table~\ref{tbl:crop-details} are rescaled
based on the dimensions of the feature maps. Whether these crops are in effect
is described below on a per-experiment basis. The base DNN intermediate layers
from which the MCs extract features are described in
Section~\ref{sec:design-mc-inputs}.

\subsection{Defining Event F1 Score}
\label{sec:eval-f1}

Most classification metrics operate on a per-frame basis. Because \FF is
event-centric, we adopt a modified recall metric from recent work that is
designed for events that span multiple frames~\cite{lee:sysml2018}. For an event
$i$, the resulting $\textit{EventRecall}_i$ metric weighs two success measures:
$\textit{Existence}_i$ rewards detecting at least one frame from the event, and
$\textit{Overlap}_i$ rewards detecting an increasing fraction of the frames from
the event. Below, $R_i$ and $P_i$ are the ground truth and predicted event
ranges, respectively.

\setlength{\abovedisplayskip}{0pt}%
\setlength{\belowdisplayskip}{0pt}%
\begin{eqnarray*}
  \textit{Existence}_i & = & \begin{cases}
    1 & \text{if detect any frame in event } i \\
    0 & \text{otherwise}
  \end{cases} \\
  \textit{Overlap}_i & = & \sum_{j}\frac{\lvert \textit{Intersect}(R_i, P_i)
                           \rvert}{\lvert R_i \rvert} \\
  \textit{EventRecall}_i & = & \alpha\times\textit{Existence}_i +
                               \beta\times\textit{Overlap}_i
\end{eqnarray*}%

We choose $\alpha=0.9$ and $\beta=0.1$ to place greater importance on detecting
at least one frame in each event. For real-time event detection in a
surveillance setting, we believe that not missing events is more important than
capturing all frames in an event. If an application receives at least one frame
from an event, then it can demand-fetch additional frames while prioritizing
between events.

On the other hand, we retain the standard definition of \emph{precision}: the
fraction of predicted frames that are true positives (i.e.,
$\frac{\text{\# correctly detected}}{\text{total \# detected}}$). For \FF,
precision determines what fraction of bandwidth is used to send relevant frames.
A precision of 1.0 means that all bandwidth is spent sending useful true
positive frames. We combine standard precision with our modified definition of
event recall to calculate an \emph{event F1 score}---the harmonic mean of
precision and recall---which is used throughout this evaluation. An intuitive
way to conceptualize event F1 score is as a measure of end-to-end event
detection accuracy.

\subsection{Saving Wide Area Bandwidth}
\label{sec:eval-bw}

First, we demonstrate that \FF achieves its primary objective, conserving
edge-to-cloud bandwidth (Section~\ref{sec:back-chall-bw}). Specifically,
filtering on the edge with FF uses $6.3 - 13\times$ less bandwidth than heavily
compressing and uploading the full stream.

\begin{figure}[t]
  \centering

  \begin{subfigure}[c]{\columnwidth}
    \centering
    \includegraphics[width=0.9\columnwidth]{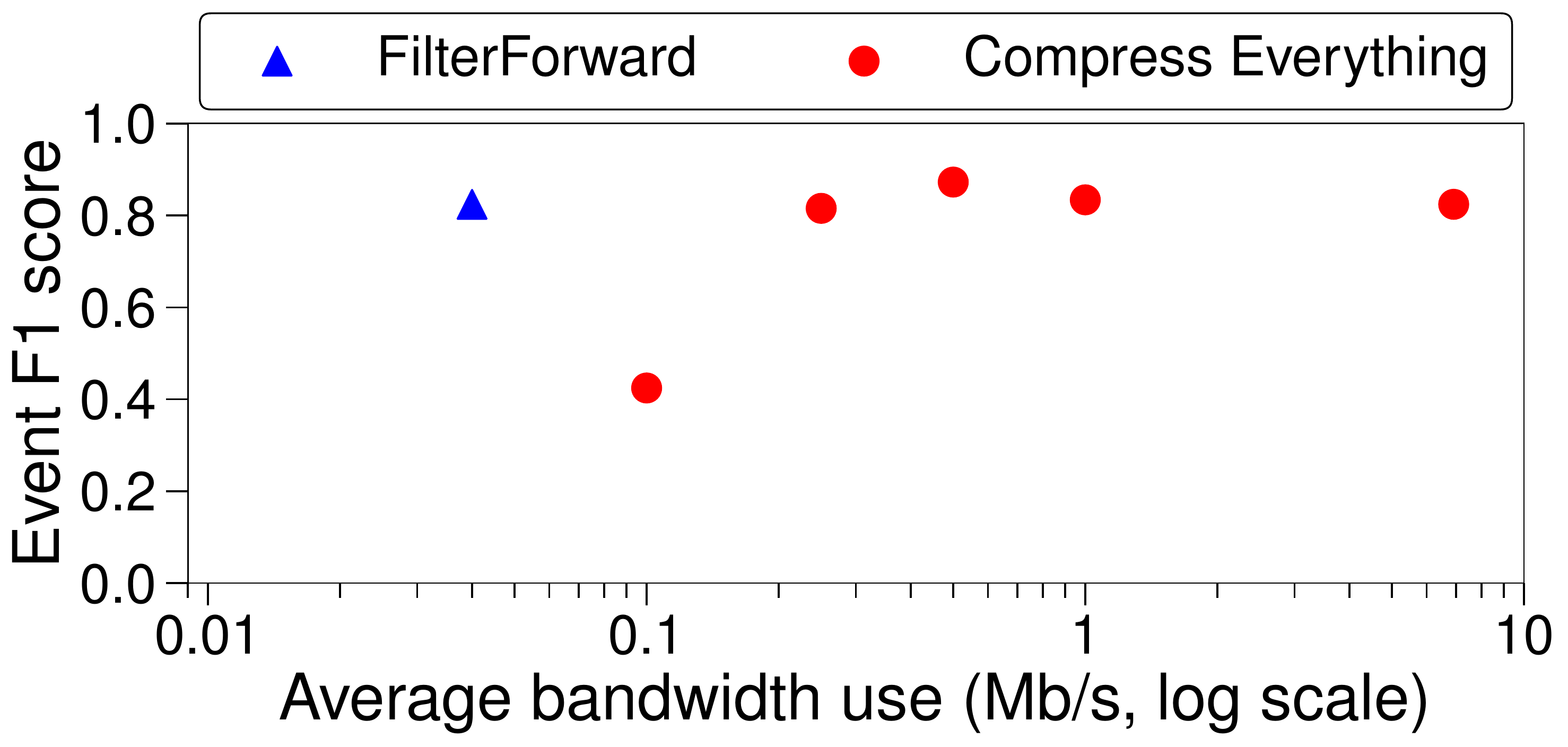}
    \caption{Full-frame object detector MC}
    \label{fig:accuracy_bandwidth_1x1_objdet}
  \end{subfigure}
  \vskip 0.15in

  \begin{subfigure}[c]{\columnwidth}
    \centering
    \includegraphics[width=0.9\columnwidth]{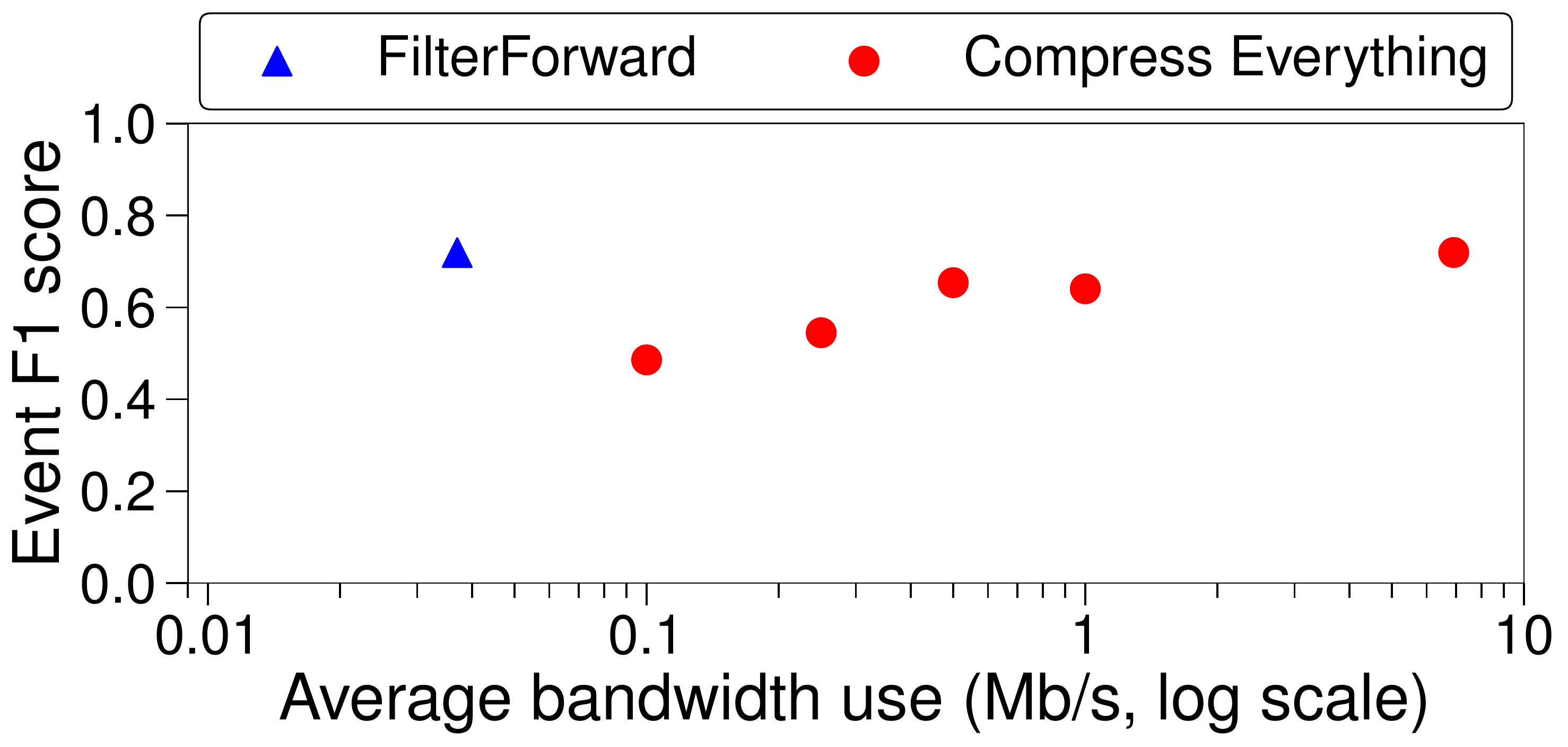}
    \caption{Localized binary classifier MC}
    \label{fig:accuracy_bandwidth_spatial_crop}
  \end{subfigure}

  \caption{Bandwidth use on the \emph{Roadway} dataset's \emph{People with red}
    task for two strategies for offloading data: (1) compressing the video using
    H.264 and sending all frames; and (2) \FF, where only relevant sequences are
    sent.}
  \vskip -1em
  \label{fig:accuracy_bandwidth}
\end{figure}

Figure~\ref{fig:accuracy_bandwidth} relates average bandwidth use and event F1
score for two MC architectures on the \emph{Roadway} dataset's \emph{People with
  red} task. In this figure, we evaluate two techniques for uploading video from
the edge: ``\FF'' corresponds to running FF on the edge, on the original stream,
and then compressing the selected frames for upload; ``Compress everything''
represents uploading the \emph{entire} stream, compressed to a low bitrate, then
running FF in the cloud. Running \FF on both the edge stream and the cloud
stream allows us to simultaneously analyze its bandwidth and accuracy
benefits. We do not evaluate other simple bandwidth-saving techniques beyond
full-stream compression because they are comparatively ineffective: (1) Reducing
the resolution is infeasible because doing so decimates small details too
aggressively; (2) Temporal sampling is nonviable because dropping a few frames
does not provide proportional bandwidth savings (video compresses well, so each
frame does not add much overhead) and arbitrarily dropping many frames can
obscure short events.

Filtering on the edge using the full-frame object detector
(Figure~\ref{fig:accuracy_bandwidth_1x1_objdet}) and localized binary classifier
(Figure~\ref{fig:accuracy_bandwidth_spatial_crop}) MCs \textbf{reduces bandwidth
  use by} $\textbf{6.3}\times$ \textbf{and} $\textbf{13}\times$, respectively,
compared to uploading the full stream. Matched frames are re-encoded to 250~Kb/s
and 500~Kb/s~\footnote{These values are used as the target bitrates for H.264
  compression. Because matched frames are bursty, the average bitrate of the
  uploaded stream is lower. I.e., there are periods where nothing is uploaded
  and then there are periods where matched frames are uploaded at these
  qualities. Furthermore, in practice, regardless of the bitrate used by the
  compression algorithm, the upload will be throttled to the maximum bandwidth
  of the network connection.}, respectively, and uploaded. These bitrates are
chosen as sufficiently good quality for the combination of task and MC. However,
it is important to note that FF's bandwidth savings is independent of the
selected upload bitrate. Whatever upload bitrate the application developer
chooses for their task, FF allows them to utilize that bandwidth more
efficiently by dropping irrelevant frames. I.e., instead of distributing the
available bandwidth uniformly across all frames, FF allows the user to
concentrate their limited bandwidth resources on the frames that matter most,
thus delivering those frames to the datacenter at the highest possible quality.

Of course, the number of frames that FF drops depends on the rarity of the
events that the MCs are searching for. The lower the aggregate detection
frequency, the more bandwidth FF will save. In
Figure~\ref{fig:accuracy_bandwidth}, the localized binary classifier MC saves
more bandwidth (i.e., drops more frames) than the full-frame object detector MC
because it experiences more false negatives (i.e., it misses some events).

In terms of accuracy, compared to heavily compressing the full stream to a
bandwidth similar to that used by FF, the full-frame object detector and
localized binary classifier MCs \textbf{increase the event F1 score by}
$\textbf{1.5}\times$ \textbf{and} $\textbf{1.9}\times$, respectively. This
demonstrates the value of processing high-fidelity data: When using similar
amounts of bandwidth, FF achieves much higher accuracy than uploading the full
stream because filtering the original data on the edge gives FF access to
fine-grained details that compression destroys. In effect, FF combines the
accuracy of sending the original video with the bandwidth savings of heavy
compression.

\subsection{End-to-end Performance Scalability}
\label{sec:eval-e2e}

\FF embraces performance scalability as a first-class design objective
(Section~\ref{sec:back-chall-scalability}). To demonstrate microclassifiers' low
marginal cost, we compare to two alternative filtering techniques: (1) na\"ively
running multiple instances of a full DNN (MobileNet), and (2) training
specialized pixel-level classifiers. The specialized classifiers, referred to as
\emph{discrete classifiers (DCs)} because they process raw pixels, are similar
to techniques used in NoScope~\cite{kang:vldb2017} (discussed further in
Section~\ref{sec:rel-comp}). A DC is faster than a general-purpose image
classification DNN like MobileNet but more expensive than an
MC. Section~\ref{sec:eval-mc} offers a more detailed cost and accuracy
comparison with DCs. \FF, the full DNNs, and the DCs all operate on full
resolution frames, which for these experiments are $1920\times1080$~pixels.

We constructed several DCs with between 100~million and 2.5~billion
multiply-adds, varying the number of convolutional layers ($2 - 4$), the number
of kernels ($16 - 64$), the stride length ($1 - 3$), the number of pooling
layers ($0 - 2$), and the type of convolutions (standard or separable). We fixed
the kernel size to 3. We report results for a representative example from the
Pareto frontier of accuracy and cost.

All performance experiments are conducted on a desktop computer with a quad-core
Intel\textsuperscript{\textregistered{}} Core\texttrademark{} i7-6700K CPU and
32~GB of RAM, using only the CPU (not the integrated or discrete GPUs). In our
experience, this CPU is representative of an edge node mounted on a light post,
but we expect future deployments to also contain GPUs or DNN hardware
accelerators. We execute the base DNN using a version of the Caffe deep learning
framework~\cite{www-caffe} that has been optimized for Intel
CPUs~\cite{www-intel-caffe} and uses the Intel Math Kernel Library for Deep
Neural Networks~\cite{www-intel-mkl-dnn}. We execute the MCs and DCs using
TensorFlow~\cite{Abadi:osdi2016}. We set the neural network batch size
independently for the full DNNs, DCs, and MCs based on a short parameter
sweep. To reduce CPU contention, we use end-to-end flow control to guarantee
that, for \FF, the base DNN and MCs are executed in phases (not pipelined) so
that Caffe and TensorFlow do not compete for cores. Evaluation videos are
H.264-compressed, reside on disk, and are always written back to disk (to
simulate archiving the full stream for lazy upload) so all performance
experiments implicitly contain disk reads/writes and H.264 decoding. At no time
are disk accesses or decoding the bottleneck. Because our testbed software stack
is not heavily optimized, the magnitude of our performance measurements matters
less than the trends in how the different architectures scale.

\begin{figure}[t]
  \centering \includegraphics[width=0.9\columnwidth]{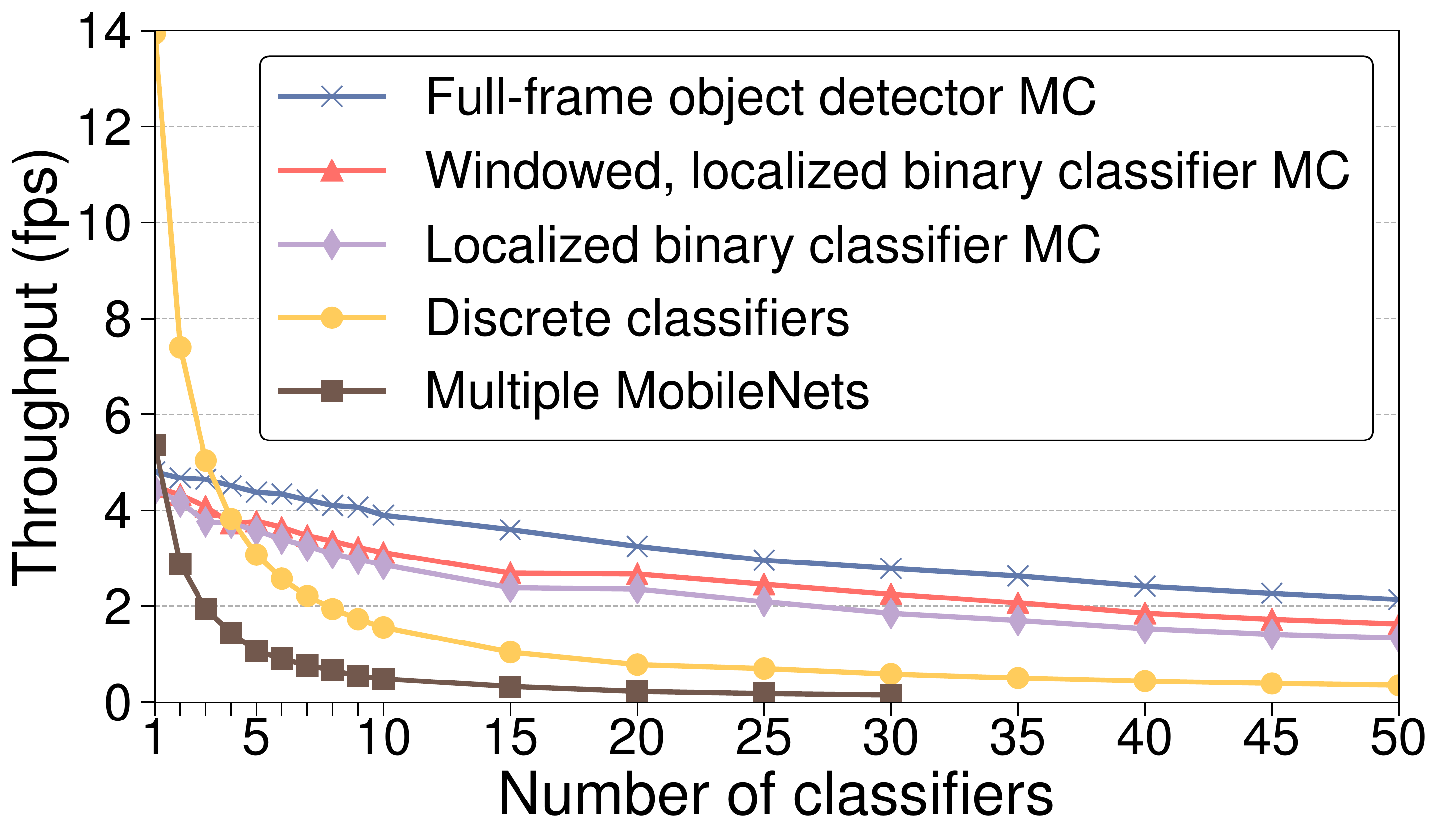}
  \caption{Throughput (in frames per second) of the three MC architectures
    compared to full DNNs and discrete classifiers. \FF amortizes the cost of
    the base DNN when running 4 or more concurrent MCs.}
  \vskip -1em
  \label{fig:e2e_tput}
\end{figure}

Figure~\ref{fig:e2e_tput} compares the filtering throughput of \FF's three MC
architectures to that of multiple full MobileNet DNNs and NoScope-style discrete
classifiers. With only a single classifier, \FF processes frames at
$0.32 - 0.34\times$ the speed of the DCs and $0.83 - 0.90\times$ the speed of
multiple MobileNets. By 20 classifiers, this has risen to $3.0 - 4.1\times$
faster than the DCs. By 50 classifiers, \FF has up to \textbf{6.1$\times$}
\textbf{higher throughput}. Intuitively, with only a single filtering task,
using MobileNet directly yields higher throughput than \FF because of feature
extraction and data movement overheads in the latter. By two classifiers, these
overheads do not matter. On the other hand, the DCs have higher throughput when
the number of classifiers is low because they do not pay the overhead of running
an expensive full DNN. However, since each DC must compute the full translation
from pixels to a decision, they perform redundant work. By sharing compute and
amortizing the cost of its base DNN, \FF is \textbf{faster with more than}
$\textbf{3 - 4}$ \textbf{classifiers}. Running multiple MobileNets, while
straightforward, is never optimal from a throughput perspective and runs out of
memory beyond 30 classifiers. Ultimately, for use cases with more than a handful
of filtering tasks, \FF offers much higher throughput than existing techniques.

\begin{figure*}[t]
  \centering

  \begin{subfigure}[c]{0.32\textwidth}
    \centering
    \includegraphics[width=\textwidth]{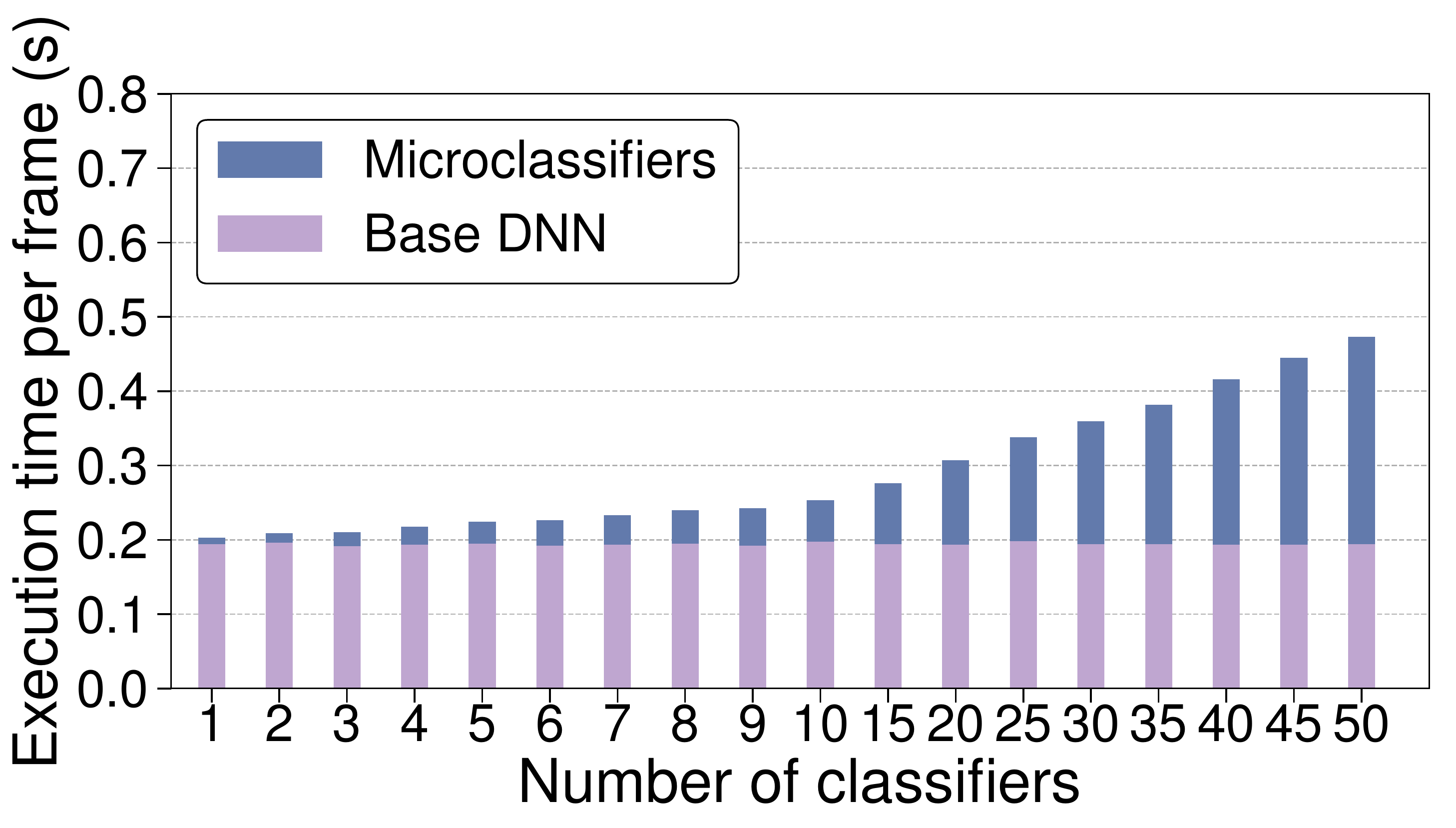}
    \caption{Full-frame binary classifier}
    \label{fig:e2e_1x1_objdet_cpu_breakdown}
  \end{subfigure}\hfill
  ~
  \begin{subfigure}[c]{0.32\textwidth}
    \centering
    \includegraphics[width=\textwidth]{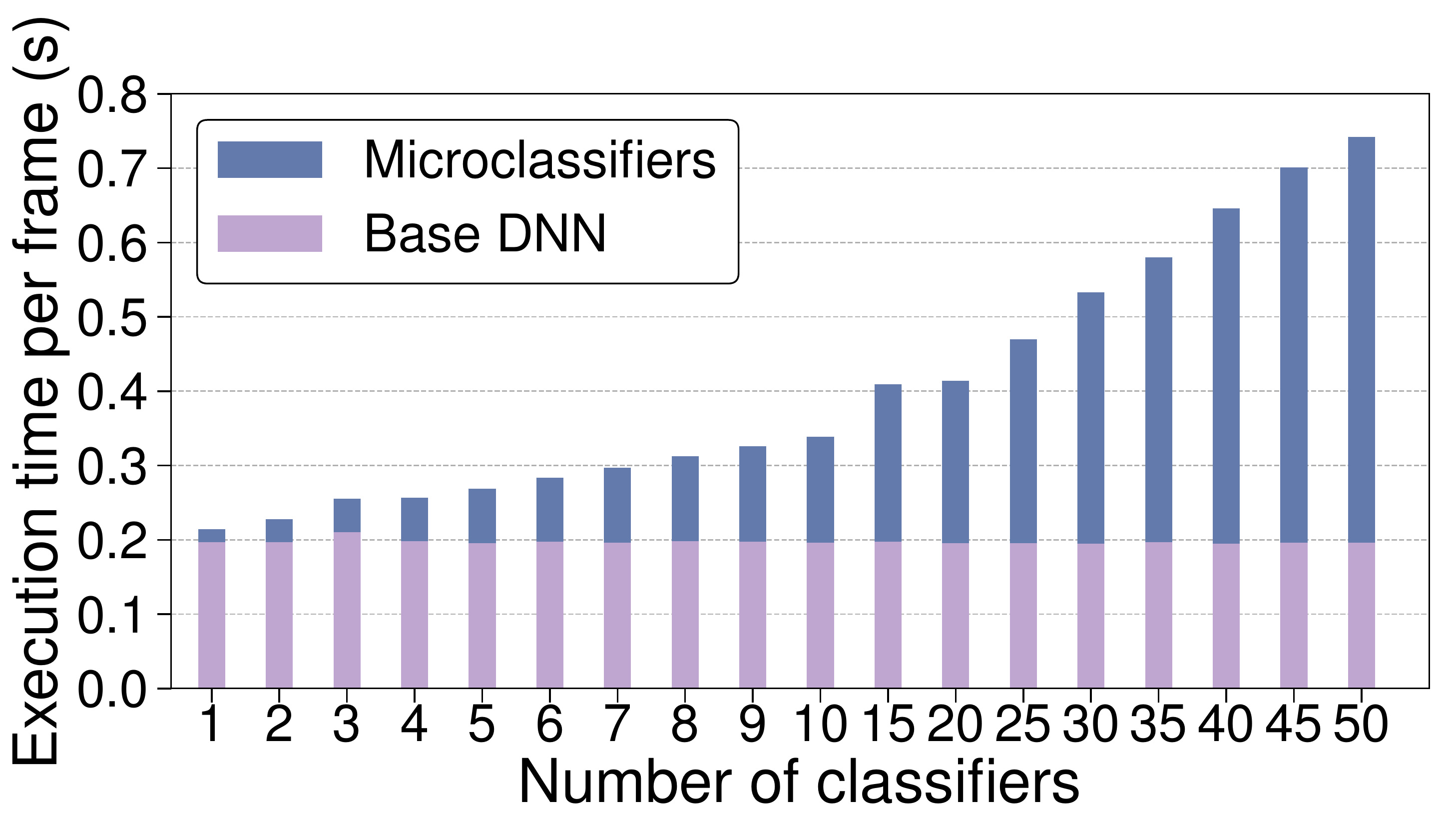}
    \caption{Localized binary classifier}
    \label{fig:e2e_spatial_crop_cpu_breakdown}
  \end{subfigure}\hfill
  ~
  \begin{subfigure}[c]{0.32\textwidth}
    \centering
    \includegraphics[width=\textwidth]{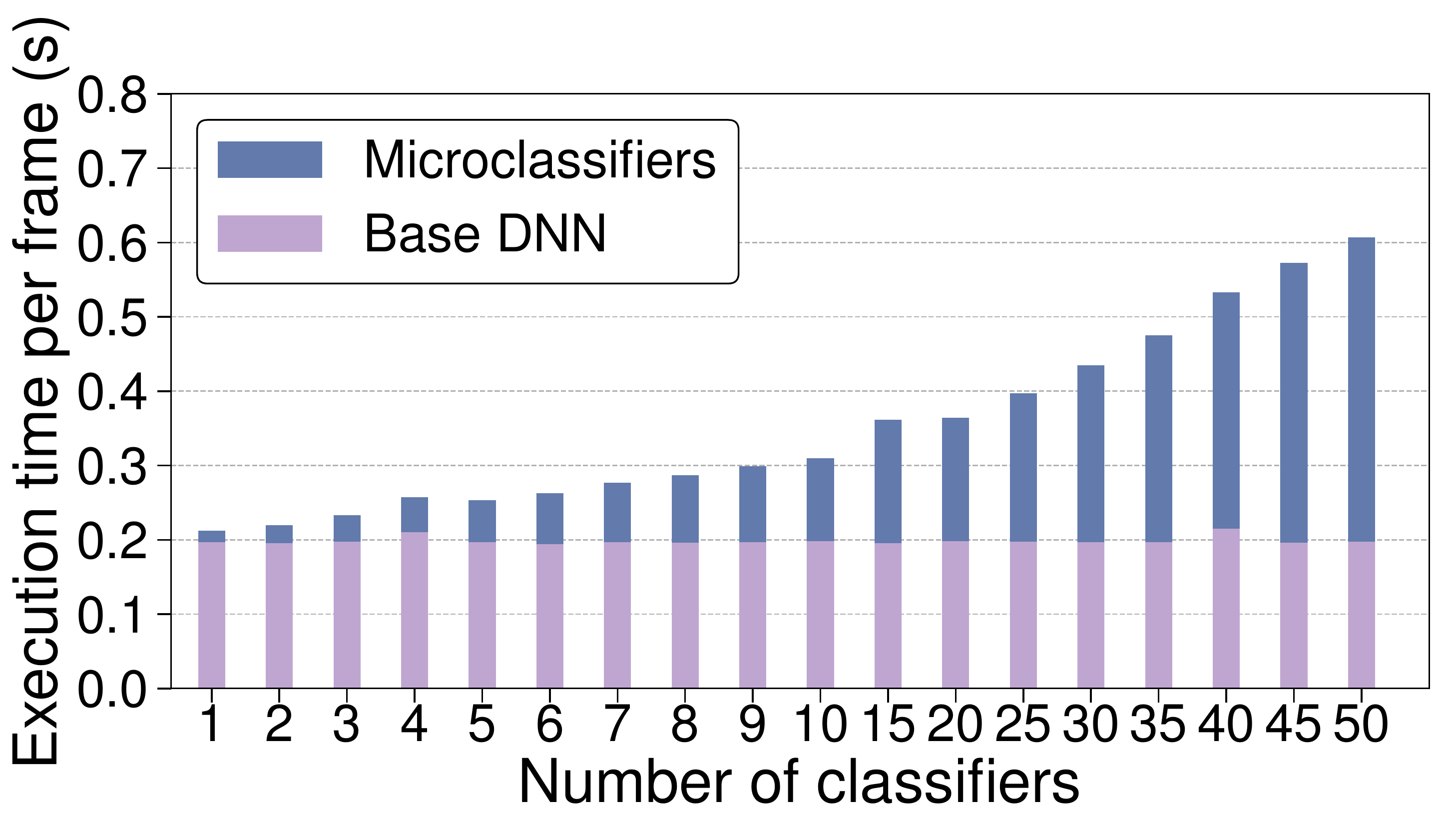}
    \caption{Windowed, localized binary classifier}
    \label{fig:e2e_windowed_cpu_breakdown}
  \end{subfigure}

  \caption{Execution time (in seconds) breakdown of \FF's main components for
    the three microclassifier architectures. \FF pays the upfront cost of
    evaluating the base DNN, but then reaps the resulting benefit of each
    additional MC being cheap.}
  \vskip -1em
  \label{fig:e2e_cpu_breakdown}
\end{figure*}

To further understand \FF's throughput scalability, for each frame we measure
the time taken by the base DNN and MCs. Figure~\ref{fig:e2e_cpu_breakdown} shows
this breakdown for our three proposed microclassifier architectures. With few
queries, the base DNN's execution time dominates, as expected. The total
execution time grows only modestly as we add dozens of concurrent MCs.
Depending on the microclassifier, the base DNN's CPU time is equivalent to that
of 15 - 40 MCs.

\subsection{Microclassifier Cost and Accuracy}
\label{sec:eval-mc}

Finally, we demonstrate that because they operate on feature maps, \FF's
microclassifiers have substantially lower marginal compute cost, yet higher
accuracy on real-world datasets, than discrete classifiers
(Section~\ref{sec:back-chall-real-world}).  We use the same discrete classifier
architecture as in Section~\ref{sec:eval-e2e}, and trained the MCs and DCs on
0.5 epochs of data, using spatial crops (Table~\ref{tbl:dataset-details}) for
the applicable MCs and the \emph{Roadway} dataset's DC (the \emph{Jackson}
dataset's DC's accuracy did not benefit from a spatial crop).

Multiply-adds are a good proxy for the compute cost of a DNN
model~\cite{DBLP:journals/corr/HowardZCKWWAA17}. Given a feature map of size
$H \times W$ and depth $M$, the number of multiply-adds in a fully-connected
layer with $N$ hidden units is: $N \times H \times W \times M$. For the same
feature maps, the number of multiply-adds in a convolutional layer with $F$
filters of size $K \times K$ and a stride of $S$ is:
$\frac{H}{S} \times \frac{W}{S} \times M \times K^2 \times F$. The cost of a
convolutional layer can be reduced using separable or ``factored'' convolutions
(kernels are split into depthwise followed by pointwise convolutions) with some
accuracy penalty. The number of multiply-adds in a separable convolutional layer
with the same parameters is:
$\frac{H}{S} \times \frac{W}{S} \times M \times (K^2 + F)$. Recall that \FF
ingests full resolution video, so in our experiments, the number of
multiply-adds is much greater than for typical input sizes, such as
$224\times224$~pixels.

\begin{figure}[t]
  \centering

  \begin{subfigure}[c]{\columnwidth}
    \centering
    \includegraphics[width=0.9\columnwidth]{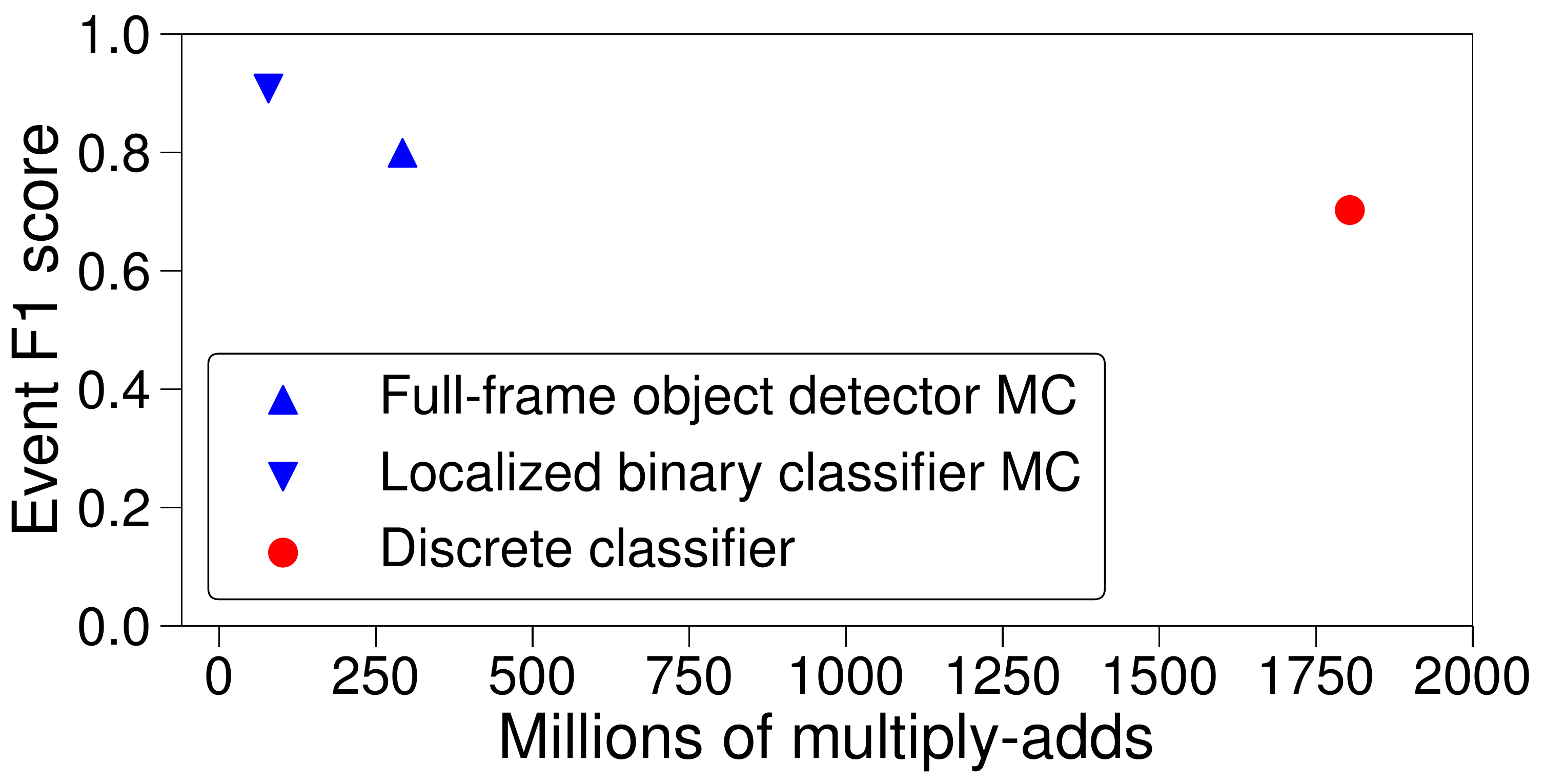}
    \caption{\emph{Jackson} dataset, \emph{Pedestrian} task.}
    \label{fig:accuracy_cost-jackson}
  \end{subfigure}
  \vskip 0.15in

  \begin{subfigure}[c]{\columnwidth}
    \centering
    \includegraphics[width=0.9\columnwidth]{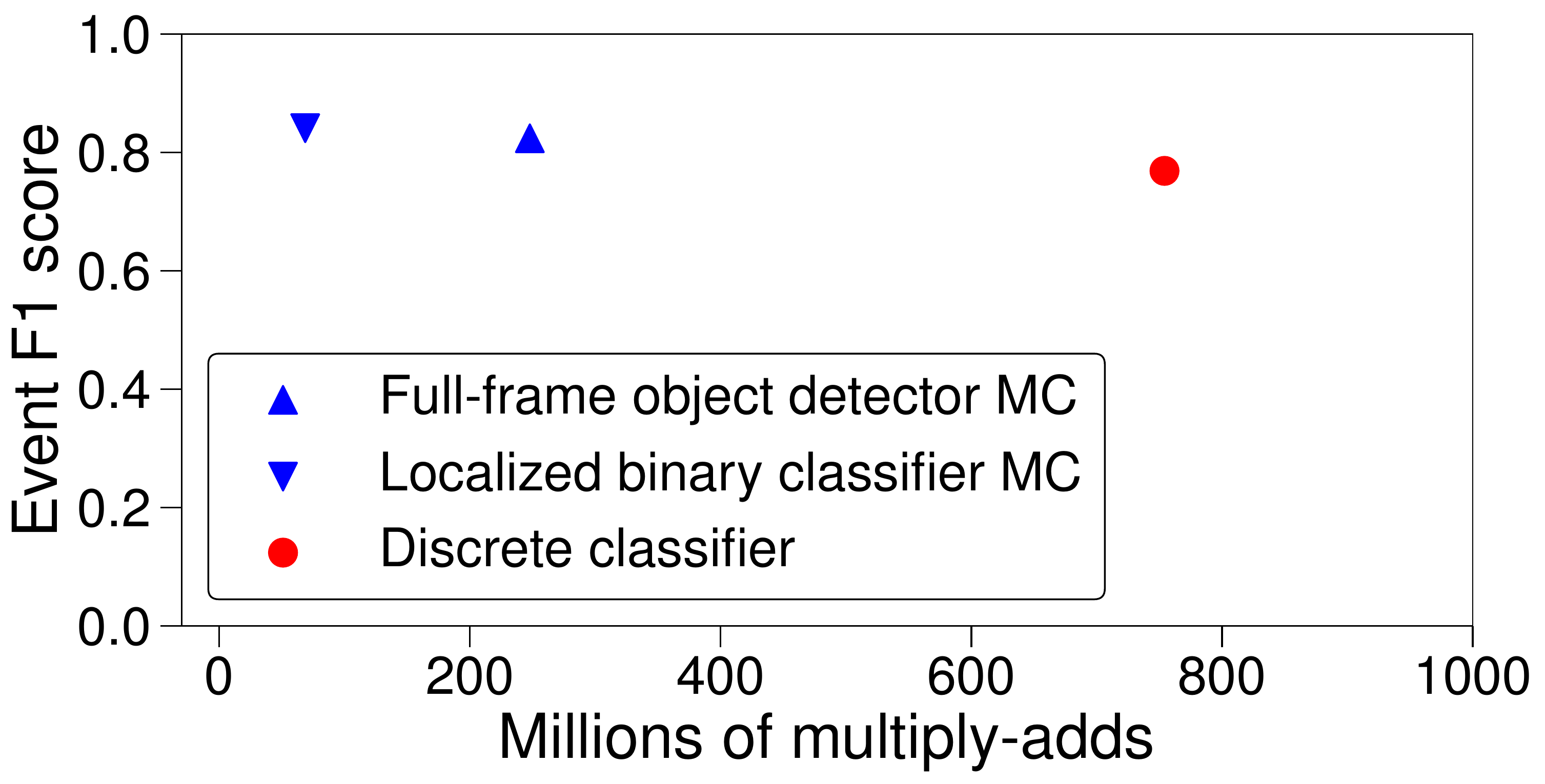}
    \caption{\emph{Roadway} dataset, \emph{People with red} task.}
    \label{fig:accuracy_cost-roadway}
  \end{subfigure}

  \caption{Number of multiply-adds versus event F1 score for microclassifiers
    and discrete classifiers. MCs have a much lower marginal cost than DCs, yet
    achieve higher accuracy.}
  \vskip -1em
  \label{fig:accuracy_cost}
\end{figure}

Figure~\ref{fig:accuracy_cost} compares two microclassifiers' accuracy (event F1
score) and marginal compute cost (number of multiply-adds) to those of the
discrete classifiers for both of our datasets. Compared to the DCs, \FF's MCs
are up to $1.3\times$ as accurate while being $23\times$ cheaper on the
\emph{Jackson} dataset and $1.1\times$ as accurate and $11\times$ cheaper on the
\emph{Roadway} dataset. This order of magnitude savings in marginal compute cost
results from the MCs operating on feature maps instead of pixels, meaning that
they have less work to do (translating features to a classification is simpler
than doing so for pixels). Of course, the tradeoff is that the MCs must pay the
upfront overhead of the base DNN.

We believe that, in Figure~\ref{fig:accuracy_cost}, the slight accuracy
improvements compared to the DCs are a byproduct of using a more complex network
for the pixel processing. The DCs must walk a fine line between accuracy and
cost: To remain lightweight compared to full multi-class DNNs like MobileNet,
they sacrifice complexity. By amortizing the pixel processing across all its
MCs, FF allows users to run a more powerful feature extraction network, and
therefore extract higher-quality features that are a better basis for additional
analysis, than would be possible using the DCs.


\section{Related Work}
\label{sec:rel}

This section outlines related work that applies to the three challenges raised
in Section~\ref{sec:back-chall-chall} (limited bandwidth, real-world video
streams, and scalable multi-tenancy)

\subsection{Conventional Machine Learning Approaches}
\label{sec:rel-ml}

Conventional ML techniques for reusing computation improve scalability, but
their rigidity sacrifices accuracy.

Transfer learning accelerates multi-application training and inference by
leveraging the observation that DNNs trained for image classification and object
detection identify general features that transfer well to specialized
tasks~\cite{donahue2014decaf,yosinski2014transferable}. During inference,
transfer learning shares computation by running one base DNN to completion and
extracting its last layer's activations as a feature vector, which is then used
by multiple specialized classifiers (one per
application)~\cite{pakha:hotcloud2018}. Recent work allows application-specific
DNNs to share multiple layers with the base DNN~\cite{Jiang:2018}, similar to
how our microclassifiers can pull from any layer of \FF's base DNN. However,
conventional transfer learning suffers from poor accuracy for small objects
because it retains the original DNN architecture for the retrained
layer(s). Even though these approaches are computationally efficient, they are
not tailored to real-world video streams.

Multi-task learning~\cite{caruana1998multitask} offers an efficient way to share
computation across models, but all models must be retrained when new tasks are
added. This retraining overhead makes multi-task learning unsuited to real-world
deployments, where tasks are frequently added and removed.

\subsection{Filtering-based Approaches}
\label{sec:rel-filtering}

Filtering video by dropping irrelevant frames reduces computation and
transmission load~\cite{kang:vldb2017,pakha:hotcloud2018,wang:sec2018}.  One
method of filtering is to use a cascade of progressively more accurate and
expensive detectors, stopping execution at the cheapest model that produces a
high confidence prediction.  This is a common technique for optimizing the
``fast path'' where most frames can be discarded near the beginning of the
cascade. Early work in this field includes \cite{viola:cvpr2001}, which
introduces a detector cascade based on traditional computer vision features and
includes an attention mechanism to prune the feature space and improve
throughput. \FF builds on this idea but specializes to the task of detecting
small objects in surveillance video.  Similar to the aforementioned attention
mechanism, FF includes an optional optimization where a microclassifier can
spatially crop its feature map to focus on a particular region (to improve
accuracy) and reduce model complexity (to save computation).

\subsubsection{Saving Compute During Bulk Analytics}
\label{sec:rel-comp}

Recent work has applied filter cascades to reduce computation load during bulk
video analytics. NoScope~\cite{kang:vldb2017} drops frames whose pixel-level
differences from a reference image or previous frame do not meet a threshold,
before feeding them into the cascade. NoScope first evaluates cheap,
task-specific, pixel-level CNNs (e.g., a custom ``Shetland pony'' binary
classifier), which we refer to as \emph{discrete classifiers}, and only applies
an expensive CNN (e.g., YOLO9000~\cite{DBLP:journals/corr/RedmonF16}) when the
confidence of the cheap CNN is below a threshold. Discrete classifiers are
similar to our MCs, except that they operate on raw pixels. In \FF, the base DNN
amortizes pixel processing across all MCs, reducing the marginal cost of each
classifier without sacrificing accuracy. We compare the throughput and accuracy
of our MCs to NoScope's discrete classifiers in Sections~\ref{sec:eval-e2e} and
\ref{sec:eval-mc}.

Previous systems were often evaluated on highly-curated datasets, where video
processing was orchestrated to be easier. For example,
NoScope~\cite{kang:vldb2017} was evaluated on video that has been cropped to a
narrow region of interest (objects typically occupy the majority of the
frame). Enlarging objects in this way makes analysis both easier (because
objects are more prominent) and cheaper (because the DNN input resolution can be
reduced). However, modifying the data in this way diverges from our goal of
processing wide-angle surveillance video. \FF supports a similar cropping
technique, but this is not crucial to its design.

Focus~\cite{hsieh:osdi2018} divides processing between ingest time and query
time, using cheap CNNs and clustering to build an approximate index up front
that dramatically accelerates offline queries. Focus's ingest CNN is
conceptually similar to \FF's base DNN---they both generate semantic information
about each frame that is used for future processing---but our use of feature
maps instead of top classes is more general. The notion of storing per-frame
metadata in an index is applicable to \FF and an interesting direction for
future work.

Both NoScope and Focus assume that it is possible to stream all video to a
resource-rich datacenter. This is not fundamental to their algorithms, but
pushing components of either system to the edge would introduce additional
compute constraints. A basic premise of \FF is that uploading all video is
infeasible, so our design builds off computation sharing that enables an edge
node to support many concurrent applications.

\subsubsection{Saving Bandwidth on Constrained Edge Nodes}
\label{sec:rel-bw}

Similar to \FF, others have approached the challenges of running ML workloads on
edge-generated video in real time. Both \cite{pakha:hotcloud2018} and
\cite{wang:sec2018} push computation to the edge to determine which frames are
``uninteresting'' to heavyweight analytics in the cloud.

\cite{pakha:hotcloud2018} uses sampling and superposition coding to send frames
only when relevant objects appear and then using the lowest possible
quality. While the work displays impressive bandwidth savings, the iterative
communication between the edge and the cloud limits its throughput.

\cite{wang:sec2018} examines the heavily bandwidth-constrained use case of
offloading video in real time from a swarm of autonomous drones using the 4G LTE
cellular network. Similar to \FF, this system uses lightweight DNNs (e.g.,
MobileNet) running on the edge (here, on the drones) combined with lightweight
classifiers (they use support-vector machines (SVMs)) to give an early
indication of whether a frame is interesting. These SVMs are similar in
principle to our microclassifiers, but always operate on activations extracted
from the base DNN's final pooling layer and are much shallower than MCs, meaning
that they have a lower capacity to learn and inferior accuracy.

Additionally, both of these systems are not optimized for multi-tenant
environments. \FF is designed with query scalability as a first-class concern,
and can run dozens of concurrent microclassifiers.

Both \cite{pakha:hotcloud2018} and \cite{wang:sec2018} focus on streams where
the camera is moving, whereas \FF considers stationary surveillance cameras.
Operating on streams with less global motion gives \FF an advantage because it
is easier to train classifiers for these streams, and the larger proportion of
unchanging pixels makes such streams more compressible.

\subsection{Resource Scheduling for Video Pipelines}
\label{sec:rel-sched}

Resource management is crucial for practical video analytics because
applications often impose the conflicting goals of maximizing their overall
benefit and meeting performance constraints. For instance,
VideoStorm~\cite{Zhang:nsdi2017} adjusts query quality to maximize a combined
utility, using efficient scheduling that leverages offline quality and resource
profiles. LAVEA~\cite{Yi:icdcs2017} places tasks across edge nodes and clients
(e.g., mobile phones) to minimize the latency of video analytics.
DeepDecision~\cite{Ran:infocom2018} expresses resource scheduling in video
processing as a combinatorial optimization problem.
Chameleon~\cite{jiang:sigcomm2018} dynamically adjusts a video processing
pipeline's hyperparameters as the content in the scene changes, using temporal
and spatial (i.e., across nearby cameras) correlations to prune the optimization
search space.

Much of this scheduling work is complementary to \FF, which shares a similar
motivation of balancing accuracy and throughput, but focuses on edge nodes with
constrained network bandwidth. Unlike prior scheduling work that adjusts only
general knobs such as video bitrate, resolution, and choice of DNN model, \FF's
computation sharing directly improves the computational efficiency of multiple
filters running on the same edge node.


\section{Conclusion}
\label{sec:concl}

Scaling real-time, wide-area video analytics poses a challenge for
bandwidth-limited, compute-constrained camera deployments. This paper presents
\FF, a new filtering architecture for the edge that uses lightweight,
per-application microclassifiers to identify relevant video segments to
upload. We show that FF reduces bandwidth use by an order of magnitude without
sacrificing accuracy while scaling to as much as $6.1\times$ higher throughput
than existing approaches. However, even though this paper describes \FF in terms
of saving bandwidth on the edge, we believe that our scalable early-discard
algorithm is a viable method of eliding unnecessary computation in ML-based
cloud analytics as well. We believe that \FF's computation sharing and hybrid
edge-to-cloud design transcend video processing and provide useful building
blocks for ML applications in constrained environments.

\FF is open source at
\href{http://www.github.com/viscloud/ff}{github.com/viscloud/ff}.


\section*{Acknowledgments}
\label{sec:acks}

We appreciate the insights of the SysML `19 program committee and our colleagues
at Carnegie Mellon University and Intel Labs. This work was supported by Intel
via the Intel Science and Technology Center for Visual Cloud Systems (ISTC-VCS).

\bibliography{ms}
\bibliographystyle{sysml2019}

\end{document}